# Some Considerations and a Benchmark Related to the CNF Property of the Koczy-Hirota Fuzzy Rule Interpolation

Maen Alzubi[#], Szilveszter Kovacs[#]

[#] *Department of Information Technology, University of Miskolc, H-3515 Miskolc, Hungary*
*E-mail: alzubi@iit.uni-miskolc.hu, szkovacs@iit.uni-miskolc.hu*

*Abstract*— The goal of this paper is twofold. Once to highlight some basic problematic properties of the KH Fuzzy Rule Interpolation through examples, secondly to set up a brief Benchmark set of Examples, which is suitable for testing other Fuzzy Rule Interpolation (FRI) methods against these ill conditions. Fuzzy Rule Interpolation methods were originally proposed to handle the situation of missing fuzzy rules (sparse rule-bases) and to reduce the decision complexity. Fuzzy Rule Interpolation is an important technique for implementing inference with sparse fuzzy rule-bases. Even if a given observation has no overlap with the antecedent of any rule from the rule-base, FRI may still conclude a conclusion. The first FRI method was the Koczy and Hirota proposed "Linear Interpolation", which was later renamed to "KH Fuzzy Interpolation" by the followers. There are several conditions and criteria have been suggested for unifying the common requirements an FRI methods have to satisfy. One of the most common one is the demand for a convex and normal fuzzy (CNF) conclusion, if all the rule antecedents and consequents are CNF sets. The KH FRI is the one, which cannot fulfill this condition. This paper is focusing on the conditions, where the KH FRI fails the demand for the CNF conclusion. By setting up some CNF rule examples, the paper also defines a Benchmark, in which other FRI methods can be tested if they can produce CNF conclusion where the KH FRI fails.

*Keywords*— Sparse Fuzzy Rule-Bases; Fuzzy Interpolation Techniques; KH FRI; CNF Property; Normality Benchmark Examples

## I. Introduction

Fuzzy inference systems that are based on classical fuzzy inference approaches have been widely applied for many real-world applications. These methods require dense rule-bases, i.e. the universe of discourse demands to be fully covered by the rule antecedent fuzzy sets [1]-[4]. The size of a dense fuzzy rule-base is usually growing exponentially, as the number of input variables (antecedent dimension) increases. This behavior, the exponentially growing rule base size and the related computational time restricts the practical implementation of the classical fuzzy inference approaches in high dimensional applications [5].

One possible solution is the decrease of the rule-base size by permitting sparse fuzzy rule-bases. The sparse or incomplete rule-bases cannot fully cover the universe of discourse. Hence classical fuzzy inference approaches fail for some fuzzy observations. The sparse fuzzy rule-base, as a knowledge representation could be a result of an intentional designer decision, or simply a sign of insufficient knowledge during the rule-base generation. This situation can be easily handled by Fuzzy Rule Interpolating (FRI) methods, where the rule matching reasoning concept is replaced by FRI function.

Fuzzy Rule Interpolation (FRI) concept was presented to provide reasonable and meaningful conclusion even if the available rules (in case of sparse rule-bases) do not match the current observation. Many FRI techniques suggested during the past two decades. The first one was proposed by Koczy and Hirota (KH FRI) [6]-[11] which was originally referred as "linear interpolation" by the authors. Then the KH FRI became the base for many other interpolation techniques. The KH FRI has many limitations. It is restricted to convex and normal antecedents and consequents fuzzy sets, having bounded support. It also demands at least a partial ordering between the fuzzy sets in the universes of discourse. A modification of the KH FRI is called VKK method which was proposed by Vass, Kalmar and Koczy [12], this method is based on α-cut technique, where the conclusion is computed based on the distance of the center points and the widths of the α-cuts, instead of lower and upper distances.

The stabilized KH (KHstab) technique was proposed by Tikk et. [13] to handle and exclude the abnormality, where this method will be used by the inverse of the distance between antecedents and observation, where all flanking of the current observation will be used in computing the conclusion. Another modification of original KH method is the Modified α-Cut based Interpolation (MACI) method which was proposed by Tikk and Baranyi [14], MACI converts fuzzy sets to vector description, then calculates the conclusion, and finally, it converts back to the initial space. Another fuzzy interpolation technique was proposed by Koczy et al. [10]. It is called conservation of relative fuzziness (CRF) method. This method was proposed to modify the fuzziness term and to improve α-cut levels, which follows fundamental equation (FEFRI) (Definition 3), where the conclusion can be obtained by determining the core and fuzziness of antecedents, consequents and observation fuzzy sets.

The original KH FRI produces the output based on its α-cuts. The most significant benefit of the KH FRI is its low computational complexity that guarantees reasoning speed required by real-time applications. Despite many advantages, in some antecedent fuzzy set configuration, the KH FRI suffers from abnormality of the conclusion (see more details in [15], [24]). The study in [16], [17] discuss the normality property and gives some boundary conditions for the observation, the antecedent and consequent fuzzy sets, where the normality of the conclusion necessarily holds.

The main goal of this paper is to take these boundary conditions and construct some Benchmark Examples to highlight the problematic properties of the original KH

Fuzzy Rule Interpolation. Besides, this Benchmark Examples could be used for testing other FRI methods against these ill conditions. All Benchmark Examples introduced in this paper are implemented by the MATLAB FRI Toolbox [18], [19] which provides an easy-to-use framework for FRI applications.

The rest of the paper is organized as follows: Section (II) provides a brief review of the basic definitions of the fuzzy set and interpolative reasoning concept. Section (II.A) introduces a summary of the original KH FRI. Overview of the main equations and notations for the normality property of the KH FRI as shown in section (II.B) and the reference values (corollaries) of the CNF property in section (II.C). A CNF Benchmark Examples of implemented KH FRI presented in section (II.D). Section (III) presents the results discussion of the Benchmark Examples, and comparing selected FRI methods based on Benchmark Examples is presented in section (III.A). Finally, section (IV) is dedicated to the conclusion of the paper.

## II. MATERIAL AND METHOD

This section a brief review of the basic definitions of the fuzzy set and interpolative reasoning concept. Typically, the fuzzy set theory is an extension of the classical set theory, it provides a general procedure for extending crisp domains of mathematical expressions to fuzzy domains, and generalizes a normal point-to-point mapping of a function to a mapping between fuzzy sets. Since can be considered specific cases of fuzzy sets, classical sets properties are extended. The degree of membership of an element of a fuzzy set evaluates by the unit interval [0, 1]. Then a fuzzy set A of the universe of discourse X is represented by its membership function $\mu A(x) \in [0,1]$, $x \in X$.

A fuzzy set defined on a universe of discourse which holds total ordering is a convex and normal fuzzy (CNF) set, if it has a height equal to one, and having membership grade of any elements between two other elements greater than, or equal to the minimum membership degree of these tow boundary elements. I.e a convex fuzzy set can be defined by $\forall x, y \in U$, $\forall \lambda \in [0,1]$: $\mu A(\lambda x + (1-\lambda y)) \geq \min(\mu A(x), \mu A(y))$.

Fig.1 describes some properties of the membership functions. The support of a fuzzy set is the set of all elements in the universe of discourse with membership degree is greater than zero. It can be defined by Supp (A): $x \in U$, $\mu A(x) > 0$. The α-cut, and the strong α-cut of a fuzzy set is the crisp subset of the universe where the membership degrees are greater (strong α-cut), or greater, or equal (α-cut) than a specified α value. The α-cut can be represented by $A\alpha$ : $x \in U$, $\mu A(x) \geq \alpha$, $\alpha \in [0,1]$ and $A_\alpha = x \in U$, $\mu A(x) > \alpha$, $\alpha \in [0,1]$. The kernel of a fuzzy set is the crisp subset of the universe where the membership degrees are equal to 1. Kernel(A): $x \in U$, $\mu A(x) = 1$.

The width of a convex fuzzy set is the length of the support, which in the case of a convex fuzzy set is an interval. The width of a convex fuzzy set is defined by Width(A): max(Supp(A)) – min (Supp(A)). The height of a fuzzy set is the maximum membership degree of all the elements of the universe, and it can be defined by Height(A): $\max(x) \in U(\mu A(x))$. A fuzzy set is said to be normal if at least one element of the universe has a membership degree equal to 1, $\exists x \in U$, $\mu A(x)$: Height(A) = 1.

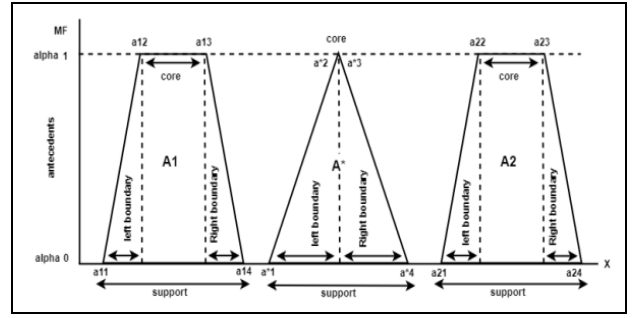

Fig.1: Support and α-Cuts of Triangular and Trapezoidal Fuzzy Sets

A typical example of KH FRI based reasoning in sparse rule-base systems is presented in [20]. It could be briefly described as follows. From the rule-base the two closest surrounding fuzzy rules to observation are taken into consideration only (see Fig.1 the observation and the two surrounding antecedent fuzzy sets):

If X is $A_1$ then Y is $B_1$
If X is $A_2$ then Y is $B_2$

The two rules will be abbreviated as $A_1 \Rightarrow B_1$ and $A_2 \Rightarrow B_2$ respectively. Suppose that these two rules are adjacent as shown in Fig.2. Thus, we can see that when observation $A^*$ has no overlapping with fuzzy sets $A_1$ or $A_2$, therefore none of the rules are firing, no results could be obtained by classical fuzzy reasoning.

If X is $A_1$ then Y is $B_1$
If X is $A_2$ then Y is $B_2$
Observation: X is $A^*$
----------------------------------
Conclusion: Y = ($B^*$)

FRI reasoning methods could provide interpolated conclusion $B^*$ when the observation $A^*$ is not overlapping with any of the rule antecedents $A_1$ and $A_2$. According to the interpolation concept which was suggested by Koczy and Hirota in [6], [7] and [21], some definitions can be introduced as follows:

**Definition 1**. Referring to all fuzzy sets that must be normal and convex in the universe $X_i$ by $P(X_i)$. The α-cuts are intervals. Then for $A_1$, $A_1 \in P(X_i)$, if $\forall \alpha \in (0,1]$, $A_1$ is precedes $A_2$ ($A_1 \prec A_2$) if:

$$\inf(A_{1\alpha}) < \inf(A_{2\alpha}), \sup(A_{1\alpha}) < \sup(A_{2\alpha}) \quad (1)$$

where $A_{1\alpha}$ and $A_{2\alpha}$ are α-cut sets of $A_1$ and $A_2$, respectively, inf ($A_{i\alpha}$) is the infimum of $A_{i\alpha}$ and sup ($A_{i\alpha}$) is the supremum of $A_{i\alpha}$ (i= 1,2).

**Definition 2**. Given a fuzzy relation $R\prec$: ($A_1$, $A_2$) | $A_1$, $A_2$ $\in P(X)$, $A_1 \prec A_2$, if fuzzy sets $A_1$ and $A_2$ satisfy $R\prec$, the lower (dL) and the upper (dU) fuzzy distances between $A_1$ and $A_2$ by using the resolution principles can be defined [6], [7], [8] as follows:

$$dL (A_1, A_2): R\prec \rightarrow P([0,1])$$

µdL(δ): ∑$_{α∈[0,1]}$ α /d(inf($A_{1α}$), inf($A_{2α}$))

dU ($A_1$, $A_2$): R≺ → P([0,1])

µdU(δ): ∑$_{α∈[01]}$ α /d(sup($A_{1α}$), sup($A_{2α}$))

where δ ∈ [0,1] and d is the Euclidean distance, or more generally the Minkowski distance.

**Definition 3.** When $A_1 ⇒ B_1$, $A_2 ⇒ B_2$ be disjoint fuzzy rules on the universe of discourse X x Y, and $A_1$, $A_2$ and $B_1$, $B_2$ be fuzzy sets on X and Y, respectively. Assume that $A^*$ is the observation of the input universe X. If $A_1 < A^* < A_2$ then the KH linear fuzzy rule interpolation between $R_1$ and $R_2$ is defined as follows:

$$d(A_1, A^*) : d(A^*, A_2) = d(B_1, B^*) : d(B^*, B_2) \quad (2)$$

where d refers to the fuzzy distance according to Definition 2 that could be used between the fuzzy sets ($A_1$, $A^*$, $A_2$) and ($B_1$, $B_2$).

### A. The Original KH Fuzzy Linear Interpolation

The original Koczy and Hirota interpolation (later referred as KH FRI) [6]-[11] requires the antecedents and consequences fuzzy sets to be convex and normal (CNF) [22], [23]. This case the approximated conclusion can be generated by decomposing the fuzzy sets into α-cuts. The KH FRI is defined for a single dimensional antecedent space, for two rules, whose antecedents are surrounding the observation:

$$A_1 ≺ A^* ≺ A_2$$
$$\text{And}$$
$$B_1 ≺ B_2$$

According to the concept of fuzzy distance [8] appearing in the KH FRI (see Definition 2), the fuzzy distance of two CNF sets can be defined as the distance of lower and upper endpoints of their α-cuts. The "linear interpolation" idea of the KH FRI is that the rate of the upper and lower fuzzy distances between observation and antecedents must be the same as the rate of the fuzzy distances between the two rule conclusions and the consequent. Therefore, regarding the previous definitions and resolution principles of fuzzy sets, the conclusion $B^*$ for the KH FRI method are produced directly based on α-cuts of the observation and the two surrounding fuzzy rules.

According to equations that introduced in [6]-[8] the conclusion of the KH FRI could be calculated as follows:

The right core of the conclusion:

$$RCB^* = \frac{d_1RC \times RCB_1 + d_2RC \times RCB_2}{d_1RC + d_2RC} \quad (3)$$

where

$$d_1RC = \sqrt{\sum_{i=1}^{k}(RCA_i^* - RCA_{i1})^2}$$

$$d_2RC = \sqrt{\sum_{i=1}^{k}(RCA_{i2} - RCA_i^*)^2}$$

The right end of the support of the conclusion:

$$RFB^* = \frac{d_1RF \times RFB_1 + d_2RF \times RFB_2}{d_1RF + d_2RF} \quad (4)$$

where

$$d_1RF = \sqrt{\sum_{i=1}^{k}(RFA_i^* - RFA_{i1})^2}$$

$$d_2RF = \sqrt{\sum_{i=1}^{k}(RFA_{i2} - RFA_i^*)^2}$$

For calculating the left-core ($LCB^*$) and the left-support ($LFB^*$) of the conclusion similar as Equations 3 and 4 could be constructed.

A key advantage of the original KH approach is its low computational complexity for fuzzy rules since it deals with two rules only from the rule base during the determination of consequent, where the antecedents of those rules are the closest flanking to the observation, $A_1 ≺ A^* ≺ A_2$ (See Fig.2).

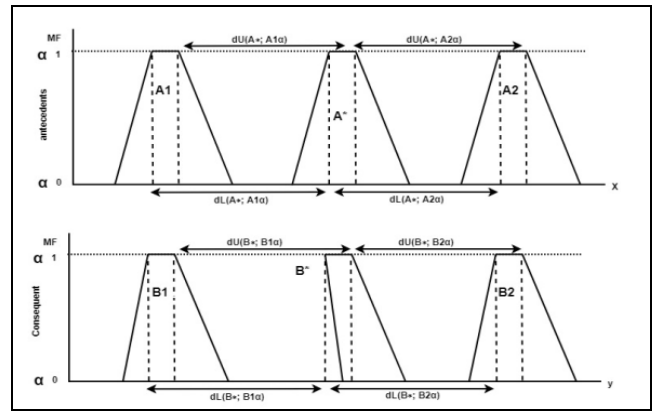

Fig.2: Fuzzy Interpolative Reasoning with an Invalid Conclusion for The KH FRI

On the other hand, for searching for these two rules could be a computationally demanding task. Despite the advantages, in some rule and observation configurations, the conclusion can be abnormal, or not always directly interpretable. Therefore, in the following section, the conditions of the normal, or abnormal conclusion will be studied in details.

### B. The Convexity and Normality of The KH FRI Conclusion

Several FRI techniques followed the resolution principle which requires turning the problem of fuzzy interpolation into an infinite family of crisp interpolations according to the α-cuts of the rules and observation, then merging the results and conclude the fuzzy solution, as follows:

$$B^* = \bigcup_{\alpha \in [0,1]} \alpha . B_\alpha^*$$

Moreover, several necessary conditions could hold. In case of the KH FRI, it is required, that all fuzzy sets must be convex and normal (CNF). This condition guarantees that all α-cuts are intervals and exists.

The CNF property of the conclusion fuzzy set can be checked if all α-cuts are connected. The KH FRI cannot produce any results, if the α-cuts are not connected, which means the KH FRI must preserve the PWL for α ∈ [0, 1] (see cases in [25]). Hence, the conclusion is created as intervals by determining their lowest and highest endpoints. Then, the convexity condition is automatically satisfied. Fig.3. represents a convex and a non-convex fuzzy set.

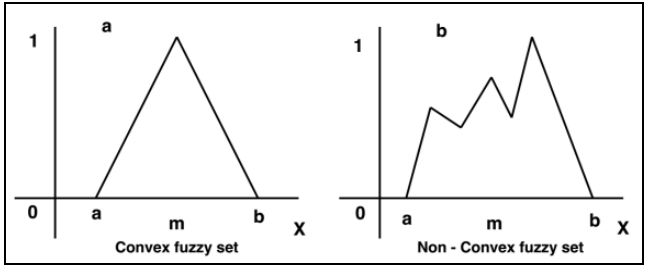

Fig.3: A Convex (A) and a Non-Convex (B) Fuzzy Set

In contrast, the normality of the conclusion is not always satisfied. The conclusion is normal, if the membership function assumes all values between 0 and 1. So, the condition will be satisfied if $\inf(B^*_\alpha) \leq \sup(B^*_\alpha)$ for all α. Otherwise, if the condition is not satisfied, the membership function will suffer from an abnormality as shown in Fig.4.

Now, we should collect which equations in [16], [17] have been used to determine and verify the normality of conclusion. In case if the shape of the antecedent and consequent fuzzy sets are restricted to triangular and trapezoidal, the membership functions can be described by three, or four points. In case of trapezoidal, it has four values ($a_1, a_2, a_3, a_4$) and in case of triangular, it could consider as a special trapezoidal $a_2 = a_3$ (see Fig.1).

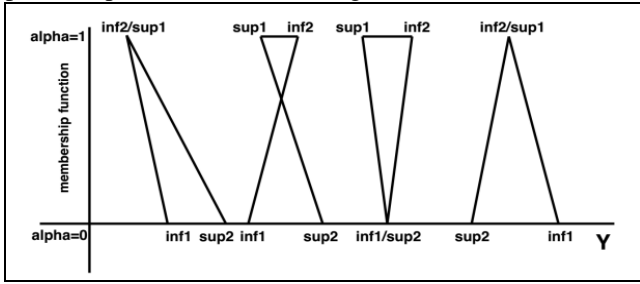

Fig.4: Forms of The Abnormal Conclusions

Additionally, a singleton membership function is also a special trapezoidal membership function where all the values ($a_1, a_2, a_3, a_4$) are the same. Accordingly, the characteristic points of the KH FRI conclusion can be defined by the following equations in [17]. Where the conclusion $B^*$ is normal if and only if $y_{inf1}$, $y_{inf2}$, $y_{sup1}$ and $y_{sup2}$ are met the following conditions:

$$y_{inf1} = \frac{(a_{21}-a_1^*)*b_{11}+(a_1^*-a_{11})*b_{21}}{a_{21}-a_{11}} \leq$$
$$y_{inf2} = \frac{(a_{22}-a_2^*)*b_{12}+(a_2^*-a_{12})*b_{22}}{a_{22}-a_{12}} \leq$$
$$y_{sup1} = \frac{(a_{23}-a_3^*)*b_{13}+(a_3^*-a_{13})*b_{23}}{a_{23}-a_{13}} \leq$$
$$y_{sup2} = \frac{(a_{24}-a_4^*)*b_4+(a_4^*-a_{14})*b_{24}}{a_{24}-a_{14}}$$

According to Equations 5-7, the core and boundary lengths of the conclusion can be determined. For verifying the normality of the LefT Boundary (LTB) length of the conclusion, Equation 5 could be applied:

$$\text{Length.LT Bound1} \leq \text{Length.LT Bound2} \quad (5)$$

where

Length.LT Bound1= $db_{LTB} \times$
$(((Ka_{1LTB}+da_{1LTB}) \times (Ka_{2LTB}+da_{2LTB})) -$
$((Ka^*_{LTB}+da_{1LTB}) \times (Ka^*_{LTB}+da_{2LTB})))$

Length.LT Bound2=$((Ka_{1LTB}+da_{1LTB}) \times$
$(da_{1LTB}+Ka^*_{LTB}) \times Kb_{2LTB})+$
$((Ka_{2LTB}+da_{2LTB}) \times$
$(da_{2LTB}+Ka^*_{LTB}) \times Kb_{1LTB})$

The core length of the conclusion can be determined by Equation 6 as follows:

$$\text{Length.Core1} \leq \text{Length.Core2} \quad (6)$$

where

Length.Core1=$db_{core} \times$
$(((Ka_{1core}+da_{1core}) \times (Ka_{2core}+da_{2core})) -$
$((Ka^*_{core}+da_{1core}) \times (Ka^*_{core}+da_{2core})))$

Length.Core2=$((Ka_{1core}+da_{1core}) \times$
$(da_{1core}+Ka^*_{core}) \times Kb_{2core}) +$
$((Ka_{2core} + da_{2core}) \times$
$(da_{2core} + Ka^*_{core}) \times Kb_{1core})$

For RighT Boundary (RTB) length of the conclusion can be determined by the following Equation 7:

$$\text{Length.RT Bound1} \leq \text{Length.RT Bound2} \quad (7)$$

where

Length.RT Bound1= $db_{RTB} \times$
$(((Ka_{1RTB}+da_{1RTB}) \times (Ka_{2RTB}+da_{2RTB})) -$
$((Ka^*_{RTB}+da_{1RTB}) \times (Ka^*_{RTB}+da_{2RTB})))$

Length.RT Bound2=$((Ka_{1RTB}+da_{1RTB}) \times$
$(da_{1RTB}+Ka^*_{RTB}) \times Kb_{2RTB})+$
$((Ka_{2RTB}+da_{2RTB}) \times$
$(da_{2RTB}+Ka^*_{RTB}) \times Kb_{1RTB})$

Where the parameters of the core length for Equation 6 can be defined as follows:

$Ka_{1core}=a_{13}−a_{12}$, $Ka_{2core}=a_{23}−a_{22}$
$Kb_{1core}=b_{13}−b_{12}$, $Kb_{2core}=b_{23}−b_{22}$
$Ka^*_{core}=x_3−x_2$, $da_{1core}=x_2−a_{13}$

Similarly to the core length parameters, the left and right equations of LTB and RTB can be constructed.

Moreover, from another point of view, the length ratio of the distance between the fuzzy sets of the antecedent with observation ($Ka_i$, $Ka^*$) and consequent ($Kb_i$) Equations 8-10 could also be used to check the normality (validity) of the conclusion, which can be defined as follows:

For the length ratio of the left Boundary:

$$RatioLT1= LTBound\ (K_{b1,b2})\ /\ LTBound\ (K_{a1,a2}).$$
$$RatioLT2= LTBound(K_{a1,a2})\ /\ (LTBound\ (Ka^*Ka_1) + LTBound(Ka_2Ka^*)). \qquad (8)$$

where

$LTBound(K_{b1,b2})= b_{21} - b_{12}$,
$LTBound(K_{a1,a2})= a_{21} - a_{12}$,
$LTBound(Ka^*Ka_1)= a^*_1 - a_{12}$,
$LTBound(Ka_2Ka^*)= a_{21} - a^*_2$.

For the length ratio of the core:

$$RatioC1= Core(K_{b1,b2})/Core(K_{a1,a2}),$$
$$RatioC2=Core(K_{a1,a2})/(Core(Ka^*Ka_1) +Core(Ka_2Ka^*)). \qquad (9)$$

where

$Core(K_{b1,b2}) = b_{22} – b_{13}$,
$Core(K_{a1,a2}) = a_{22} – a_{13}$,
$Core(Ka^*Ka_1) = a^*_2 - a_{13}$,
$Core(Ka_2Ka^*) = a_{22} - a^*_3$.

For the length ratio of the right Boundary:

$$RatioRT1= RTBound\ (K_{b1,b2})\ /\ RTBound\ (K_{a1,a2}).$$
$$RatioLT2= RTBound(K_{a1,a2})\ /\ (RTBound\ (Ka*Ka1) + RTBound(Ka_2Ka^*)). \qquad (10)$$

where

$RTBound(K_{b1,b2})= b_{23} – b_{14}$,
$RTBound(K_{a1,a2})= a_{23} – a_{14}$,
$RTBound(Ka^*Ka_1)= a^*_3 – a_{14}$,
$RTBound(Ka_2Ka^*)= a_{23} - a^*_4$.

### C. Reference Values for The CNF Property

According to the main corollaries in [16], [17], the normality of the KH FRI conclusion can be determined as follows:

*1) Corollary 1:* when ($Ka_i$ = $Kb_i$ = $Ka^*$). If rules $A_1 \Rightarrow B_1$, $A_2 \Rightarrow B_2$ and the observation $A^*$ have the same core and left-right boundary lengths as the antecedent ($Ka_i$) and consequent ($Kb_i$) fuzzy sets, the conclusion will always be normal. For this corollary, Equations 5-7 and Equations 8-10 could be used to validate the normality.

*2) Corollary 2:* when ($Ka_i$ = KA, $Kb_i$ = KB). If the membership functions of the antecedent ($Ka_i$ = KA), and the consequent ($Kb_i$ = KB) have uniform core and boundary lengths, then the conclusion fuzzy set is always normal if and only if the following conditions by Equations 11 and 12 are hold:

For the core length:

- If $Ka^* \neq 0$

$$Length.Core1 \leq Length.Core2 \qquad (11)$$

where

$Length.Core1= db_{core} \times (Ka_{core}−Ka^*_{core})$,
$Length.Core2= Kb \times (da_{1core}+da_{2core}+2\times Ka^*_{core})$

- If $Ka^* = 0$

$$Length.Core1 \leq Length.Core2 \qquad (12)$$

where

$Length.Core1= db_{core} \times (Ka_{core}−Ka^*_{core})$,
$Length.Core2= Kb_{core} \times da_{core}$

and

$$da_{core}=a_{22}−a_{13}$$

For left and right boundary lengths, similar equations to the core length could be constructed.

*3) Corollary 3:* when ($Ka_i$ = $Ka^*$, $Kb_i$ = KB). In this corollary, if the antecedent fuzzy sets and observation have the same core and support lengths, and the fuzzy sets of the consequent have the same length too, then the conclusion fuzzy set is always normal. To verify the normality condition, Equations 11 and 12 for the core and support lengths are used.

*4) Corollary 4:* this corollary discusses the antecedents and consequences that have uniform core length. The conclusion fuzzy set is always normal if the length ratio of the distance between the fuzzy sets of the antecedent and consequent **(distance KB) / (distance KA)** does not exceed the length ratio of themselves. Equations 8-10 can be used to verify the normality condition, in other words, the consequents have not shorter length, i.e. the consequents are not less than fuzzy the antecedents.

Fig.5 illustrates all the notations of the core and boundary (Ka, Kb) that are used in Equations 5-10, as follows:

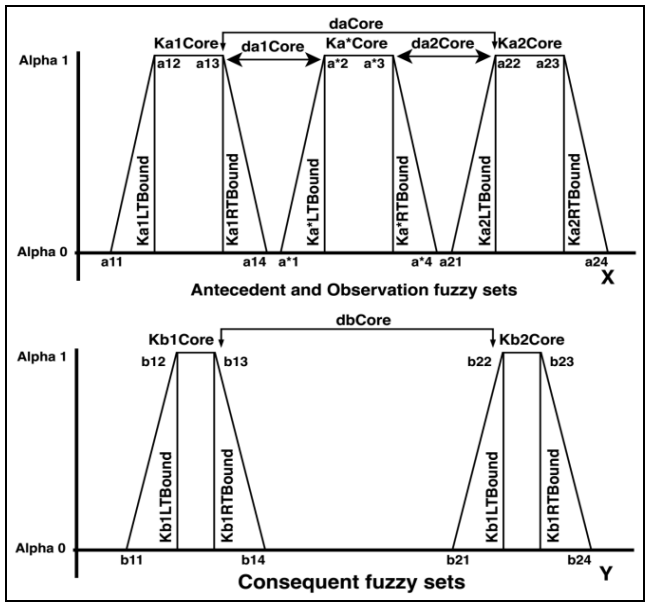

Fig.5: Notations Related to Core and Boundary Lengths of Trapezoidal Fuzzy Sets

*D. The CNF Benchmark of The KH FRI*

In the followings, Benchmark Examples will be constructed to highlight the validity of the normal property conditions of the KH FRI. Various corollaries introduced to check the normality of the KH FRI conclusion. The core and (left-right) boundary lengths have a primary role in determining the normality of the conclusion fuzzy set. According to the prerequisites of the KH FRI, one-dimensional antecedents and consequents with trapezoidal, triangular and singleton fuzzy sets, and two rules of the rule-bases could be considered. In the rest of the paper, all the calculations and figures were prepared by the fuzzy rule interpolation (FRI) toolbox. The current version of FRI toolbox is freely available to download at [18].

Now, we will discuss in details the special cases where the conclusion of the KH FRI is normal and abnormal according to the equations and corollaries that addressed previously. First of all, the normality condition is always satisfied with the KH FRI if any of the following cases are met:

- **Case 1:** when the core and boundary lengths of the observation is greater, or equal than the antecedent fuzzy sets ($KA^* >= KA$), if ($Ka_i = KA$), the normality of the KH FRI conclusion fuzzy set will always be satisfied with the normality condition. For this case, there is no restriction on the shape and size of the consequent ($KB$). Table I illustrates the Example 1 that demonstrates Case 1.

TABLE I
THE NORMAL CONCLUSION OF THE KH FRI WITH FUZZY SETS ACCORDING TO CASE 1

| Example (1) | |
|---|---|
| Notations that prove case 1 ($KA^* >= KA$), if ($Ka_i = KA$) | |
| **The values of the fuzzy sets:**<br>$A_1$=[1 2 2 3] $A_2$=[7 8 8 9]<br>$B_1$=[2 2 2 2] $B_2$=[8 8 8 8]<br>$A^*$=[4 5 5 6] $B^*$=[5 5 5 5] | **The case of the Core and (LF and RF) Boundary conclusion:**<br>The length (LFBound) is (NORMAL)<br>The length (Core) is (NORMAL)<br>The length (RFBound) is (NORMAL) |
| **The length ratio between KA, Ka* and KB:**<br>LTBound:<br>RatioLT1= (1.20), RatioLT2= (1.25)<br>Core:<br>RatioC1= (1), RatioC2= (1)<br>RTBound:<br>RatioRT1= (1.20), RatioRT2= (1.25) | **Notations length to determine the normality:**<br>LTBound1 = (0), LTBound2 = (0)<br>Core1 = (0), Core2 = (0)<br>RTBound1 = (0), RTBound2 = (0) |
| *The Antecedent (Triangular) and Observation (Triangular)* | *The Consequent (Singleton) and Conclusion (Singleton)* |

- **Case 2:** the core and boundary lengths fuzzy sets must be the same ($KA = KB$) if ($Ka_i = KA$) and ($Kb_i = KB$), the normality of the KH FRI conclusion fuzzy set will always be satisfied. For this case, there is no restriction on the shape and size of the observation $A^*$. Table II illustrates Examples 2 and 3 to demonstrate Case 2.

- **Case 3:** If the core and boundary lengths of fuzzy sets ($KB > KA$), where ($Ka_i = KA$) and ($Kb_i = KB$). The KH FRI conclusion fuzzy set will always be satisfied with normality property. Table III represents Examples 4 and 5 that describes Case 3.

TABLE II
THE NORMAL CONCLUSION OF THE KH FRI WITH FUZZY SETS ACCORDING TO CASE 2

| Example (2) | |
|---|---|
| Notations that prove case 2 when ($KA = KB$) | |
| **The values of the fuzzy sets:**<br>$A_1$=[1 2.5 2.5 4] $A_2$=[6 7.5 7.5 9]<br>$B_1$=[1 2.5 2.5 4] $B_2$=[6 7.5 7.5 9]<br>$A^*$=[4.5 5 5 5.5] $B^*$=[4.5 5 5 5.5] | **The case of the Core and (LF and RF) Boundary conclusion:**<br>The length (LFBound) is (NORMAL)<br>The length (Core) is (NORMAL)<br>The length (RFBound) is (NORMAL) |
| **The length ratio between KA, Ka∗ and KB:**<br>LTBound:<br>RatioLT1= (1), RatioLT2= (1.16)<br>Core:<br>RatioC1= (1), RatioC2= (1)<br>RTBound:<br>RatioRT1= (1), RatioRT2= (1.16) | **Notations length to determine the normality:**<br>LTBound1 = (3.5), LTBound2 = (6)<br>Core1 = (0), Core2 = (0)<br>RTBound1 = (3.5), RTBound2 = (6) |
| *The Antecedent (Triangular) and Observation (Triangular)* | *The Consequent (Triangular) and Conclusion (Triangular)* |
| Example (3) | |
| **The values of the fuzzy sets:**<br>$A_1$=[1 2 3 4] $A_2$=[6 7 8 9]<br>$B_1$=[1 2 3 4] $B_2$=[6 7 8 9]<br>$A^*$=[4 4.8 5.2 6] $B^*$=[4 4.8 5.2 6] | **The case of the Core and (LF and RF) Boundary conclusion:**<br>The length (LFBound) is (NORMAL)<br>The length (Core) is (NORMAL)<br>The length (RFBound) is (NORMAL) |
| **The length ratio between KA, Ka∗ and KB:**<br>LTBound:<br>RatioLT1= (1), RatioLT2= (1.25)<br>Core:<br>RatioC1= (1), RatioC2= (1.11)<br>RTBound:<br>RatioRT1= (1), RatioRT2= (1.25) | **Notations length to determine the normality:**<br>LTBound1 = (0.8), LTBound2 = (4.8)<br>Core1 = (2.4), Core2 = (4.4)<br>RTBound1 = (0.8), RTBound2 = (4.8) |
| *The Antecedent (Trapezoidal) and Observation (Trapezoidal)* | *The Consequent (Trapezoidal) and Conclusion (Trapezoidal)* |

TABLE III
THE NORMAL CONCLUSION OF THE KH FRI WITH FUZZY SETS ACCORDING TO CASE 3

| Example (4) | |
|---|---|
| Notations that prove case 3 when (KB > KA) | |
| **The values of the fuzzy sets:**<br>$A_1$=[1.5 2 2 2.5] $A_2$=[6.5 7 7 7.5]<br>$B_1$=[1 2 3 4] $B_2$=[6 7 8 9]<br>$A^*$=[4.5 4.5 4.5 4.5] $B^*$=[4 4.5 5.5 6] | **The case of the Core and (LF and RF) Boundary conclusion:**<br>The length (LFBound) is (NORMAL)<br>The length (Core) is (NORMAL)<br>The length (RFBound) is (NORMAL) |
| **The length ratio between KA, Ka• and KB:**<br>LTBound:<br>RatioLT1= (0.88), RatioLT2= (1)<br>Core:<br>RatioC1= (0.80), RatioC2= (1)<br>RTBound:<br>RatioRT1= (0.88), RatioRT2= (1) | **Notations length to determine the normality:**<br>LTBound1 = (2), LTBound2 = (4.5)<br>Core1 = (0), Core2 = (5)<br>RTBound1 = (2), RTBound2 = (4.5) |
| *The Antecedent (Triangular) and Observation (Singleton)* — *The Consequent (Trapezoidal) and Conclusion (Trapezoidal)* | |
| **Example (5)** | |
| **The values of the fuzzy sets:**<br>$A_1$=[2 2 2 2] $A_2$=[8 8 8 8]<br>$B_1$=[1 2 3 4] $B_2$=[6 7 8 9]<br>$A^*$=[4.5 5 5 5.5] $B^*$=[3.08 4.5 5.5 6.916] | **The case of the Core and (LF and RF) Boundary conclusion:**<br>The length (LFBound) is (NORMAL)<br>The length (Core) is (NORMAL)<br>The length (RFBound) is (NORMAL) |
| **The length ratio between KA, Ka• and KB:**<br>LTBound:<br>RatioLT1= (0.66), RatioLT2= (1.09)<br>Core:<br>RatioC1= (0.66), RatioC2= (1)<br>RTBound:<br>RatioRT1= (0.66), RatioRT2= (1.09) | **Notations length to determine the normality:**<br>LTBound1 = (-2), LTBound2 = (6.5)<br>Core1 = (0), Core2 = (6)<br>RTBound1 = (-2), RTBound2 = (6.5) |
| *The Antecedent (Singleton) and Observation (Triangular)* — *The Consequent (Trapezoidal) and Conclusion (Trapezoidal)* | |

In contrast, the abnormality of the conclusion can appear in case (KB < KA). So, to demonstrate the abnormality problem, we will consider the length ratio between $Ka^*$ and KB based on Equations 8-10. Therefore, we will address the problem with different lengths of core and boundary.

Tables IV – VII describe the results of Equations 5-10 to prove that the normality of the KH FRI conclusion will not satisfied. The Example 6 on Table IV shows the problem with the core length. Examples 7 and 8 on Tables V and VI illustrate the problem of the left and right boundary. The Example 9 on Table VII shows the problem in both the core and boundary lengths.

Regarding Table IV describes the abnormality in the core length of the KH FRI conclusion.

TABLE IV
THE PROBLEM WITH CORE LENGTH, ABNORMAL CONCLUSION

| Example (6) | |
|---|---|
| **The values of the fuzzy sets:**<br>$A_1$=[1 2 3 4] $A_2$=[6 7 8 9]<br>$B_1$=[1.5 2.5 2.5 3.8] $B_2$=[6.5 7.5 7.5 9]<br>$A^*$=[4.2 5.2 5.2 6.7] $B^*$=[4.7 5.7 4.7 6.6] | **The case of the Core and (LF and RF) Boundary conclusion:**<br>The length (LFBound) is (NORMAL)<br>The length (Core) is (**PROBLEM**)<br>The length (RFBound) is (NORMAL) |
| **The length ratio between KA, Ka• and KB:**<br>LTBound:<br>RatioLT1= (1), RatioLT2= (1.33)<br>Core:<br>RatioC1= (1.25), RatioC2= (1)<br>RTBound:<br>RatioRT1= (0.92), RatioRT2= (1.6) | **Notations length to determine the normality:**<br>LTBound1 = (0), LTBound2 = (5)<br>Core1 = (5), Core2 = (0)<br>RTBound1 = (-9.25), RTBound2 = (17.28) |
| *The Antecedent (Trapezoidal) and Observation (Triangular)* — *The Consequent (Triangular) and Conclusion (Trapezoidal)* | |

An explanation of the abnormality of the left boundary length of the KH FRI conclusion is shown on Table V.

TABLE V
THE PROBLEM WITH LEFT LENGTH, ABNORMAL CONCLUSION

| Example (7) | |
|---|---|
| **The values of the fuzzy sets:**<br>$A_1$=[1 2.5 2.5 4] $A_2$=[5.5 7.5 7.5 9]<br>$B_1$=[1 2 3 4.5] $B_2$=[6.5 7 8 9.5]<br>$A^*$=[4.5 4.9 5.1 5.5] $B^*$=[5.27 4.4 5.6 6.0] | **The case of the Core and (LF and RF) Boundary conclusion:**<br>The length (LFBound) is (**PROBLEM**)<br>The length (Core) is (NORMAL)<br>The length (RFBound) is (NORMAL) |
| **The length ratio between KA, Ka• and KB:**<br>LTBound:<br>RatioLT1= (1.5), RatioLT2= (1.15)<br>Core:<br>RatioC1= (0.8), RatioC2= (1.04)<br>RTBound:<br>RatioRT1= (1), RatioRT2= (1.12) | **Notations length to determine the normality:**<br>LTBound1 = (30.15), LTBound2 = (6.80)<br>Core1 = (-0.8), Core2 = (5.2)<br>RTBound1 = (3.85), RTBound2 = (5.85) |
| *The Antecedent (Trapezoidal) and Observation (Trapezoidal)* — *The Consequent (Trapezoidal) and Conclusion (Trapezoidal)* | |

An illustration of the abnormality in the right boundary length of the KH FRI conclusion is displayed on Table VI.

TABLE VI
THE PROBLEM WITH RIGHT LENGTH, ABNORMAL CONCLUSION

| Example (8) | |
|---|---|
| **The values of the fuzzy sets:**<br>$A_1$=[1.5 2.5 2.5 4.3] $A_2$=[6.5 7.5 7.5 8.8]<br>$B_1$=[1 2 3 3.5] $B_2$=[6 7 8 8.9]<br>$A^*$=[4.5 4.9 5.1 5.5] $B^*$=[4 4.4 5.6 4.94] | **The case of the Core and (LF and RF) Boundary conclusion:**<br>The length (LFBound) is (NORMAL)<br>The length (Core) is (NORMAL)<br>The length (RFBound) is (**PROBLEM**) |
| **The length ratio between KA, Ka• and KB:**<br>LTBound:<br>RatioLT1= (1), RatioLT2= (1.11)<br>Core:<br>RatioC1= (0.80), RatioC2= (1.04)<br>RTBound:<br>RatioRT1= (1.40), RatioRT2= (1.14) | **Notations length to determine the normality:**<br>LTBound1 = (2.4), LTBound2 = (4.4)<br>Core1 = (-0.8), Core2 = (5.2)<br>RTBound1 = (25.65), RTBound2 = (6.76) |
| *The Antecedent (Triangular) and Observation (Trapezoidal)* — *The Consequent (Trapezoidal) and Conclusion (Trapezoidal)* | |

Table VII describes the abnormality in both core and boundary lengths of the KH FRI conclusion.

TABLE VII
THE PROBLEM WITH CORE AND BOUNDARY LENGTHS, ABNORMAL CONCLUSION

| Example (9) | |
|---|---|
| **The values of the fuzzy sets:**<br>A₁=[2 2 2.5 3] A₂=[6 7.5 8 8]<br>B₁=[2 2 2 2] B₂=[8 8 8 8]<br>A*=[5 5 5 5] B*=[6.5 5.27 4.72 4.4] | **The case of the Core and (LF and RF) Boundary conclusion:**<br>The length (LFBound) is (**PROBLEM**)<br>The length (Core) is (**PROBLEM**)<br>The length (RFBound) is (**PROBLEM**) |
| **The length ratio between KA, Ka∗ and KB:**<br>LTBound:<br>RatioLT1= (1.5), RatioLT2= (1)<br>Core:<br>RatioC1= (1.2), RatioC2= (1)<br>RTBound:<br>RatioRT1= (1.2), RatioRT2= (1) | **Notations length to determine the normality:**<br>LTBound1 = (27), LTBound2 = (0)<br>Core1 = (3), Core2 = (0)<br>RTBound1 = (9), RTBound2 = (0) |

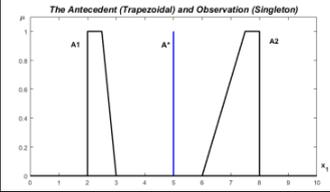

## III. RESULTS AND DISCUSSION

This section introduces the results discussion of the CNF Benchmark Examples in details. The cases and equations discussed earlier have been used to construct the Benchmark Examples. The Benchmark classified into two groups, as shown on Tables I-VII, the first one contains Examples 1-5, which regards to a normal conclusion of the KH FRI. The second includes Examples 6-9 which regards to an abnormal conclusion. For the first group, corollaries 1-4 are met, the conclusion of the KH FRI is always normal.

Referring to Example 1 on Table I, we can see that the conclusion is normal $B^* = [5\ 5\ 5\ 5]$ because the length of fuzzy sets (KA) and (Ka*) are the same, where Equations 11 and 12 are satisfying the normality conclusion. I.e. the core length Core1=0 is less than or equal to Core2=0 and for the left length LTBound1=3.5 is also less than LTBound2=6. From another side, the length ratio according to Equations 8-10 are also satisfied, i.e. for length ratio of left boundary, RatioLT1= 1.20 is less than RatioLT2= 1.25.

In Example 3 on Table II, according to case 2 the conclusion is normal $B^*=[4\ 4.8\ 5.2\ 6]$ when fuzzy sets ($Ka_1 = Ka_2$) and ($Kb_1 = Kb_2$) are the same, then Equation 11 is satisfying the normality conclusion. I.e. the right length RTBound1=0.8 is less than or equal to RTBound2=4.8. Also, according to Equations 8-10 are also satisfied for core and support, i.e. for the length ratio of the right boundary, RatioRT1=1 is less than RatioRT2= 1.25.

In another case, for Example 4 on Table III, the conclusion $B^*=[4\ 4.5\ 5.5\ 6]$ is normal as proved by Equation 12, i.e. for left and right boundary lengths, LTBound1 = 2 is less than LTBound2 = 4.5, RTBound1 = 2 is less than RTBound2 = 4.5, the conclusion is normal. Also, for length ratio by Equations 8-10, LTBound: RatioLT1= 0.88 is less than RatioLT2= 1 and RTBound: RatioRT1= 0.88 is less than RatioRT2= 1, the conclusion is normal.

Nevertheless, the second group, Examples 6-9 demonstrate some of the corollaries which are not met, when the length (KB) is less than (KA) and the length ratio (Ka* < KA). Equations 8-10 are important to prove abnormality when ratio [Core(KA)/Core(KB))] does not exceed ratio [Core(KA)/(Core(Ka*KA₁) + Core(KA₂Ka*))]. In this case, the conclusion of the KH FRI still suffers from abnormality.

In Example 6 on Table IV, the conclusion is B*=[4.7 5.7 4.7 6.6], the abnormality shown in the core length which is not fulfils (Definition 1), because the value 5.7 is greater than 4.7, in addition, notation of the length ratio of core (Equation 9) is not satisfied, as RatioC1= 1.25 exceeds the RatioC2= 1. From another side, Equation 12 also demonstrates an issue with the core length values, because Core1= 5 is greater than Core2 = 0, and therefore, the conclusion is suffering from abnormality.

For Example 7 on Table V, the conclusion is $B^*= [5.27\ 4.4\ 5.6\ 6.0]$, the abnormality be seen in left boundary length, the value 5.27 is greater than 4.4, thus Equation 8 is not satisfied, as RatioLT1= 1.5 exceeds the RatioLT2= 1.15. Also Equation 5 demonstrates a problem with the left length values, where LTBound1= 30.15 is greater than LTBound2 = 6.80, and therefore, the conclusion is suffering from abnormality.

In Example 8 on Table VI, the conclusion is B*=[4 4.4 5.6 4.94], the abnormality is shown in right boundary length because the value 5.6 is greater than 4.94 (see Definition 1). Additionally, Equation 10 of the length ratio of right boundary length is not satisfied, as RatioRT1= 1.40 exceeds the RatioRT2= 1.14. Also, Equation 7 is not fulfilled, because RTBound1= 25.65 is greater than RTBound1= 6.76.

For Example 9, the conclusion is B*=[6.5 5.27 4.72 4.4], where the problem found with core and boundary lengths. The characteristic points in the case are not matching with Definition 1, according to Equations 5-7 of the left and right boundary. Also, neither Equation 12 satisfies for the conclusion fuzzy set, as shown on Table VII:

$$LTBound1=27 > LTBound2=0,$$
$$Core1=3 > Core2=0,$$
$$RTBound1=9 > RTBound2=0.$$

In addition, Equation 8-10 are also not satisfied because (Ratio1) is greater than (Ratio2) for the core and boundary of the conclusion:

$$\text{Left ratio: RatioLT1}=1.5 > \text{RatioLT2}=1$$
$$\text{Core: RatioC1}=1.2 > \text{RatioC2}=1$$
$$\text{Right ratio: RatioRT1}=1.2 > \text{RatioRT2}=1$$

### A. Comparing Between of FRI Methods Based On Benchmark

In this section, we compare between some of the FRI methods (KHstab [13], MACI [14], VKK [12] and CRF [10]) according to the constructed Benchmark Examples 6, 7 and 9. To offer a simple way of comparison we focus on the cases that demonstrated the fails of normality of KH FRI method. This comparison shows the difference between the results of the selected methods according to the CNF property. Fig.6 introduces the antecedents part of the Examples 6, 7 and 9.

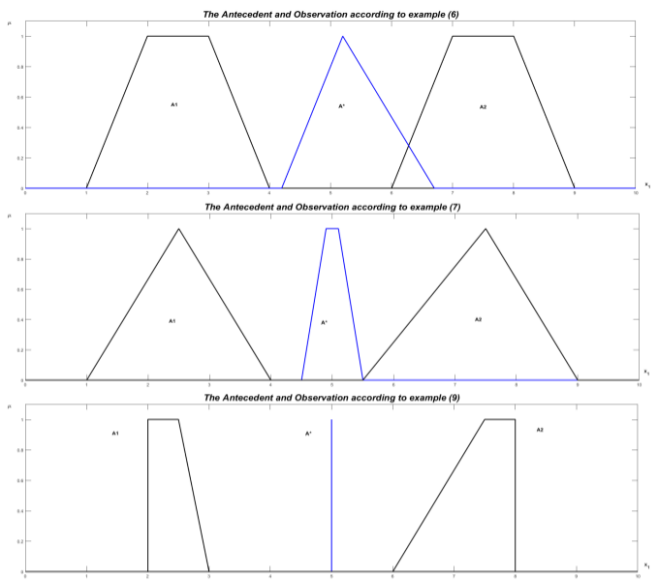

Fig.6: The Antecedents and Observations for Examples (6, 7 and 9).

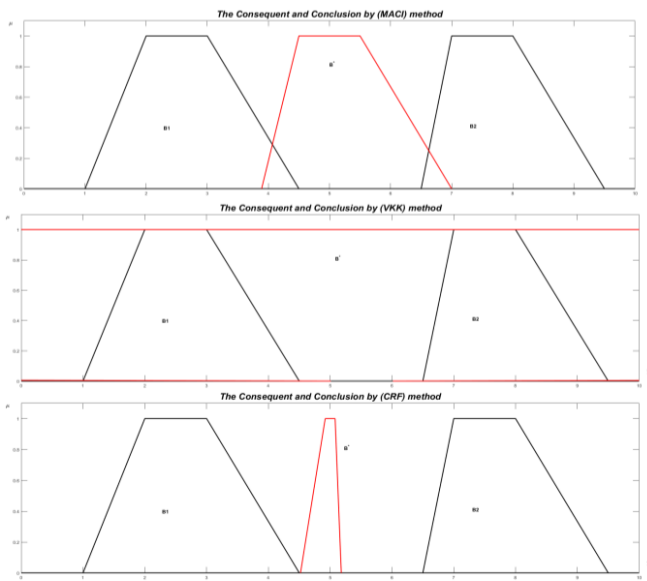

Fig.8: The Approximated Conclusion of the (KHstab [13], MACI [14], VKK [12] and CRF [10]) Methods for the Example 7.

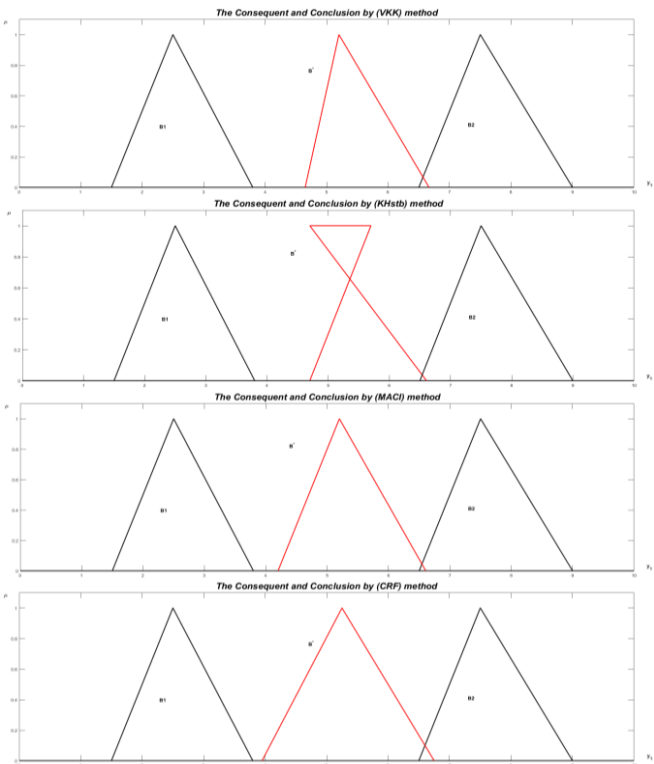

Fig.7: The Approximated Conclusion of The (KHstab [13], MACI [14], VKK [12] and CRF [10]) Methods for the Example 6.

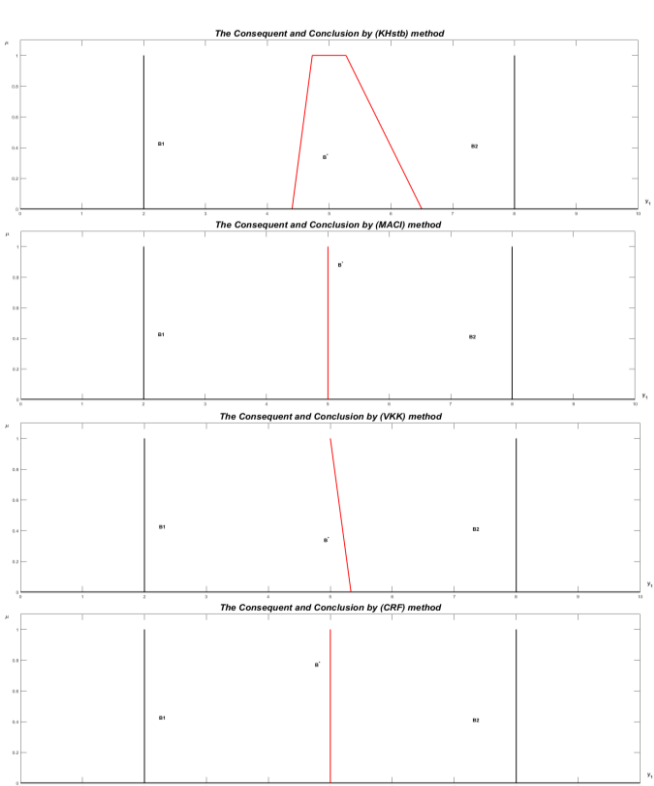

Fig.9: The Approximated Conclusion of the (KHstab [13], MACI [14], VKK [12] and CRF [10]) Methods for the Example 9.

The conclusions of the selected FRI methods (as shown in Figs.7, 8 and 9) can be discussed according to Benchmark Examples 6, 7 and 9 as follows:
- MACI [14] and CRF [10] methods are a suitable approach to be implemented as an inference system because its conclusions succeeded with CNF property to all Benchmark Examples.
- VKK [12] method, the abnormality exceeded in Benchmark Example 6 only. But it failed with CNF property in Benchmark Examples 7 and 9.

- KHstab [13] method suffered from abnormality according to all Benchmark Examples 6, 7 and 9.

Table VIIII illustrates the results of the selected FRI methods as shown by values of the conclusions B*.

TABLE IX
THE APPROXIMATE VALUES OF THE SELECTED METHODS CONCLUSIONS FOR EXAMPLES 6, 7 AND 9

| Method | Approximate conclusion B* | | |
|---|---|---|---|
| | For Example 6 | For Example 7 | For Example 9 |
| KH [6]–[11] | Abnormality [4.7 5.7 4.7 6.6] | Abnormality [5.27 4.4 5.6 6.0] | Abnormality [6.5 5.27 4.72 4.4] |
| KHstab [13] | Abnormality [4.7 5.7 4.7 6.6] | Abnormality [5.27 4.4 5.6 6.0] | Abnormality [6.5 5.27 4.72 4.4] |
| MACI [14] | Normal [4.2 5.2 5.2 6.6] | Normal [3.8 4.5 5.5 7] | Normal [5 5 5 5] |
| VKK [12] | Normal [4.6 5.2 5.2 6.66] | Abnormality [out range] | Abnormality [5.3 5 5 5.3] |
| CRF [10] | Normal [3.9 5.25 5.25 6.75] | Normal [4.5 4.9 5.0 5.1] | Normal [5 5 5 5] |

IV. CONCLUSIONS

Since the introduction of the fuzzy rule interpolation methods, several conditions and criteria have been suggested for unifying the common requirements FRI methods have to satisfy. One of the most common ones is the demand for a convex and normal fuzzy (CNF) conclusion if all the rule antecedents, consequents and the observation are CNF sets. The goal of this paper was to collect some cardinal rule-base and observation examples according to the first FRI method, the "Koczy-Hirota linear interpolation" (KH FRI) satisfies and fails the requirements for the CNF conclusion. Some corollaries and Equations 5-12 have been also set up to examine the normality of the fuzzy conclusion.

The suggested examples were constructed as a Benchmark, in which other FRI methods can be tested if they can produce CNF conclusion where the KH FRI fails.

The suggested Benchmark Examples is made up of two groups. Examples 1-5 are forming the first group in which the conclusion of the KH FRI is always normal. In opposite, Examples 6-9 are forming the second group in which the conclusion of the KH FRI is always abnormal. Examples of the first group proved the normality of the conclusion, if all cases: Case 1: ($Ka_i >= Ka^*$, Case 2 ($Ka_i < Ka^*$), and Case 3 ($Ka_i = KA$, $Kb_i = KB$) are met. While examples of the second group demonstrated the abnormality of the conclusion according to the discussed cases, when the length of KB is less than KA and the length of $Ka^*$ is less than KA. As a result, the selected FRI methods (KHstab, MACI, VKK and CRF) are compared with KH FRI based on Benchmark Examples 6, 7 and 9, which showed that the results of KHstab and VKK methods suffered from the preservation of CNF property, in contrast, MACI and CRF methods succeeded in preserving CNF property as illustrated in Table VIII.

ACKNOWLEDGMENT

The described study was carried out as part of the EFOP-3.6.1-16-00011 Younger and Renewing University - Innovative Knowledge City - institutional development of the University of Miskolc aiming at intelligent specialization project implemented in the framework of the Szechenyi 2020 program. The realization of this project is supported by the European Union, co-financed by the European Social Fund.